\documentclass[runningheads]{llncs}

\usepackage[T1]{fontenc}

\usepackage[hyphens]{url}
\usepackage{hyperref}
\usepackage{xcolor}
\usepackage[hyphenbreaks]{breakurl}

\let\llncssubparagraph\subparagraph
\let\subparagraph\paragraph
\usepackage{titlesec}
\let\subparagraph\llncssubparagraph

\usepackage[shortlabels]{enumitem}

\usepackage{placeins}

\usepackage{amsmath}

\usepackage{booktabs}
\usepackage{multirow}
\usepackage[normalem]{ulem}
\usepackage{subcaption}
\usepackage{tabu}
\usepackage{graphicx}  

\useunder{\uline}{\ul}{}

\usepackage{etoolbox}
\makeatletter
\patchcmd{\hyper@makecurrent}{%
    \ifx\Hy@param\Hy@chapterstring
        \let\Hy@param\Hy@chapapp
    \fi
}{%
    \iftoggle{inappendix}{
        \@checkappendixparam{chapter}%
        \@checkappendixparam{section}%
        \@checkappendixparam{subsection}%
        \@checkappendixparam{subsubsection}%
        \@checkappendixparam{paragraph}%
        \@checkappendixparam{subparagraph}%
    }{}%
}{}{\errmessage{failed to patch}}

\newcommand*{\@checkappendixparam}[1]{%
    \def\@checkappendixparamtmp{#1}%
    \ifx\Hy@param\@checkappendixparamtmp
        \let\Hy@param\Hy@appendixstring
    \fi
}
\makeatletter

\newtoggle{inappendix}
\togglefalse{inappendix}

\apptocmd{\appendix}{\toggletrue{inappendix}}{}{\errmessage{failed to patch}}

\newif\ifappendices

\appendicestrue

\titlespacing*{\paragraph}{0pt}{0.3\baselineskip}{\fontdimen2\font}
\titlespacing*{\subsubsection}{0pt}{0.5\baselineskip}{\fontdimen2\font}

\titlespacing*{\subsection}{0pt}{0.7\baselineskip}{0.05\baselineskip}
\titlespacing*{\section}{0pt}{\baselineskip}{0.2\baselineskip}

\begin{document}
\title{Cross-Domain Toxic Spans Detection}%
\author{%
Stefan F. Schouten\and 
Baran Barbarestani\and
Wondimagegnhue Tufa\and
Piek Vossen\and
Ilia Markov}%
\authorrunning{S. F. Schouten et al.}%
%
\institute{Vrije Universiteit Amsterdam\\
De Boelelaan 1105, 1081 HV Amsterdam, The Netherlands\\
\email{\{s.f.schouten,b.barbarestani,w.t.tufa,p.t.j.m.vossen,i.markov\}@vu.nl}\\
}%
\maketitle
\begin{abstract}
Given the dynamic nature of toxic language use, automated methods for detecting toxic spans are likely to encounter distributional shift. 
To explore this phenomenon, we evaluate three approaches for detecting toxic spans under cross-domain conditions: lexicon-based, rationale extraction, and fine-tuned language models.
Our findings indicate that a simple method using off-the-shelf lexicons performs best in the cross-domain setup.
The cross-domain error analysis suggests that (1) rationale extraction methods are prone to false negatives, while (2) language models, despite performing best for the in-domain case, recall fewer explicitly toxic words than lexicons and are prone to certain types of false positives. 
Our code is publicly available at: \url{https://github.com/sfschouten/toxic-cross-domain}.

\end{abstract}
\section{Introduction}

The rise of social media over the past decade and a half and the accompanying increase in exposure to toxic language has motivated much research into the automated detection of such language \cite{fortuna_survey_2018,pamungkas_towards_2021}.
Online toxic language use is highly dynamic and often specific to particular communities.
To deal with shifts in use of toxic language over time and to handle particular communities being underrepresented in the training data, methods for toxic language detection should generalize outside the original data distribution.
Generalization for {message-level} toxic language detection was previously investigated by evaluating methods in a cross-domain setup \cite{wiegand-etal-2019-detection}.
This has provided valuable insights into how well methods trained on data from one domain perform on data from other domains.
In this work, we investigate the 
\emph{detection of toxic spans} \cite{pavlopoulos-etal-2021-semeval} in a cross-domain setup.
In contrast to detecting overall toxicity, detecting spans aids the explainability of such systems and supports moderators in deciding on appropriate interventions sensitive to the dynamics within specific communities.

We address the following research question: how well do current methods for toxic spans detection perform in a cross-domain setting?
Our first contribution answers this question quantitatively: we evaluate three kinds of methods using the same metrics on the same datasets, reporting in-domain and cross-domain performance. 
Two experimental settings are considered: one where the overall toxicity of the texts is considered known a priori, and another where a binary toxicity classifier is used to infer the overall toxicity.
The second contribution is an in-depth error analysis of the best performing methods where we investigate and group incorrect predictions by type.

Our experimental results indicate that off-the-shelf lexicons of toxic language outperform all other methods in a cross-domain setup, whether the binary toxicity is assumed to be known or inferred.
The error analysis suggests that language models recall fewer explicitly toxic words than lexicons, and that they are prone to particular types of false positives, such as incorrectly predicting the target of the toxicity as a part of the toxic span.
\section{Related Work} \label{sec:related_work}
%
%
The task of toxic spans detection originated as a shared task at SemEval 2021 \cite{pavlopoulos-etal-2021-semeval}.
From the submissions, Pavlopoulos et al. \cite{pavlopoulos-etal-2021-semeval} identified multiple interesting approaches, three of which are described in the following paragraphs.

\textit{Lexicon-based} \label{sec:related:lexicon} approaches were widely used for message-level toxicity classification. 
They are based on word-matching techniques, which do not take context into account and miss censored or altered swear words. 
Despite this, and although these methods are unsupervised, they still achieve fairly good results \cite{fortuna_survey_2018}. 
When lexicons were used for toxic spans detection, several approaches constructed them from (span-annotated) toxic data \cite{pavlopoulos-etal-2021-semeval}.
The lexicon-based approaches performed well, with F1 scores of up to 64.98\% attained by Zhu et al. \cite{zhu-etal-2021-hitsz}.
Using a simple statistical strategy, Zhu et al. built their lexicon from the shared task's training data (see \autoref{sec:method:lexicon}). 
We include their method for constructing lexicons in our experiments and explore its effectiveness in a cross-domain setting.

\textit{Rationale extraction} \label{sec:related:sequence} techniques use 
Explainable Artificial Intelligence (XAI) methods to attribute a toxicity classifier's decision to its inputs. 
Performing the detection of toxic spans using XAI approaches assumes that the inputs that are most important to a toxicity classifier also comprise the toxic spans we aim to detect. 
A big benefit is that XAI approaches are generally unsupervised and do not require much data \cite{plucinski-klimczak-2021-ghost}. 
Different XAI methods have been used, including model-specific attention-based methods \cite{plucinski-klimczak-2021-ghost,rusert-2021-nlp}, but also model-agnostic methods such as SHAP \cite{plucinski-klimczak-2021-ghost} and LIME \cite{benlahbib-etal-2021-lisac}.
We include the rationale extraction approach in our experiments and evaluate rationales from four XAI methods under cross-domain conditions.

\textit{Fine-tuned language models} \label{sec:related:model} (LMs) formed the most popular category among the shared task submissions \cite{pavlopoulos-etal-2021-semeval}.
Both the winner and the runner-up of the shared task were based on ensembles of fine-tuned LMs \cite{zhu-etal-2021-hitsz,nguyen-etal-2021-nlp}.
Both submissions used LMs fine-tuned for sequence labeling with the BIO (Beginning, Inside, Outside) scheme, but Zhu et al. \cite{zhu-etal-2021-hitsz} also used an LM fine-tuned for span boundary detection. Others participants, such as Chhablani et al. \cite{chhablani-etal-2021-nlrg}, used models designed for extractive question answering.
We also include a fine-tuned LM in our experiments, investigating how well it performs in a cross-domain setting.

Recently, Ranasinghe \& Zampieri \cite{ranasinghe-zampieri-2021-mudes} used the dataset from the SemEval shared task to train a model with multi-lingual embeddings, evaluating on Danish and Greek datasets.
They also evaluated their model off-domain for document-level toxicity detection, whereas we evaluate cross-domain toxic span detection.
%
Previous work has also investigated \emph{message-level} toxicity classifiers under cross-domain conditions, reporting significant drops in performance \cite{pamungkas_towards_2021,markov-etal-2021-exploring,markov-daelemans-2021-improving}.
On the message-level task pre-trained language models show better generalization and ability to deal with domain shift. However, combining them with either external resources such as lexicons \cite{pamungkas_towards_2021} or with feature-engineered approaches \cite{markov-daelemans-2021-improving} can improve cross-domain prediction performance further.
Pamungkas et al. \cite{pamungkas_towards_2021} note that previous works have investigated two types of domains: topic domains (e.g., racism vs. sexism), and platform domains (e.g., Twitter vs. Facebook).
While there may be differences in the topic distributions of our domains, our primary focus is on toxic spans detection across platform domains.

To the best of our knowledge, we are the first to evaluate methods for the detection of toxic spans under cross-domain conditions. 
By doing so, we shed light on which approaches are best suited to handle shifts to out-of-domain data.

\section{Methodology}
This section describes in detail the methods for toxic spans detection we include in our experiments, and how we evaluate them.

\subsection{Evaluation}
{Our evaluation metric} is based on that used in SemEval-2021 Task 5, where Pavlopoulos et al. \cite{pavlopoulos-etal-2021-semeval}
define the following metric:
\begin{align}
    F_{1}^{+}(\mathcal{Y}, \mathcal{T}) =
    \begin{cases} 
        F_{1}(\mathcal{Y}, \mathcal{T})     & |\mathcal{T}| > 0
        \\
        1                                   & |\mathcal{T}| = |\mathcal{Y}| = 0
        \\
        0                                   & otherwise
    \end{cases}
\end{align}  
Where $\mathcal{Y}$, $\mathcal{T}$ correspond respectively to the predicted and ground truth sets of toxic character offsets, and with:
\begin{align}
F_{1}(\mathcal{Y}, \mathcal{T}) = \frac{2 \cdot P(\mathcal{Y}, \mathcal{T}) \cdot R(\mathcal{Y}, \mathcal{T})}{P(\mathcal{Y}, \mathcal{T}) + R(\mathcal{Y}, \mathcal{T})}, P(\mathcal{Y}, \mathcal{T}) = \frac{|\mathcal{Y} \cap \mathcal{T}|}{|\mathcal{Y}|},  R(\mathcal{Y}, \mathcal{T}) = \frac{|\mathcal{Y} \cap \mathcal{T}|}{|\mathcal{T}|}.
\end{align}
They introduce this modified $F_1$ score to handle texts that do not include span annotations.
We further use it to evaluate performance on non-toxic texts, which we include in our experimentation (see \autoref{sec:experiments}).
We use the same metric, but report the macro (instead of micro) average between toxic and non-toxic samples.
We do so because the chosen datasets differ in ratio of toxic to non-toxic (see \autoref{tab:dataset_balance}).
By using macro averages we can compare results across datasets. 

We investigate each method in two settings.
The first setting assumes that we know for each text if it is toxic or not, we call this setting `ToxicOracle'.
This demonstrates the ability of each method to identify toxic spans separately from their ability to identify overall toxicity.
The second setting `ToxicInferred' makes no such assumption. 
Instead, it includes a binary toxicity classifier to predict whether texts are toxic before predicting the actual toxic spans.
The errors made in the first stage are propagated to the second stage by not predicting any spans whenever the binary classifier predicts the text as non-toxic.

\subsection{Methods for Toxic Spans Detection}
We perform toxic spans detection using three distinct approaches chosen based on the results of SemEval 2021 task \cite{pavlopoulos-etal-2021-semeval}.

\subsubsection{Lexicons.} \label{sec:method:lexicon}
We use two varieties of lexicons: pre-existing lexicons of toxic language and lexicons constructed from toxic spans detection training data. 
For the latter we use the methodology proposed by Zhu et al. \cite{zhu-etal-2021-hitsz}: we quantify the toxicity of a word as the frequency with which it occurs in a toxic span relative to its overall frequency. 
The lexicon is constructed by only including words with a toxicity score higher than a certain threshold.

\subsubsection{Rationales.} 
We extract rationales from a model (in our case, BERT \cite{devlin-etal-2019-bert}) trained on binary toxicity classification (\textit{toxic} vs. \textit{non-toxic}) using various eXplainable AI (XAI) methods.
The XAI methods we use attribute the decision of a model to its inputs. 
The result is a score for each input indicating its importance relative to the other inputs. 
To obtain the toxic spans we threshold these importance scores, thereby predicting that the toxic parts of the input are those parts which were most important to the binary toxicity classifier.

\subsubsection{LMs.} 
We fine-tune an LM (BERT) for token classification using BIO labels.

\section{Experimental Details}\label{sec:experiments}

Our main experimental contribution is the systematic evaluation of methods for the prediction of toxic spans in a cross-domain setting. 
Each of our methods is evaluated both under in-domain and cross-domain conditions.

\subsection{Datasets}
Our experiments are carried out with two datasets annotated for toxic spans. 
Their similarities and differences are described below.

\paragraph{SemEval-2021 Task 5} \cite{pavlopoulos-etal-2021-semeval}. This shared task introduced a dataset of toxic samples harvested from the Civil Comments dataset, re-annotating a portion for toxic spans. 
In the campaign, annotators were asked to ``Extract the toxic word sequences (spans) of the comment [\dots], by highlighting each such span''. 
The inter-annotator agreement was ``moderate'', with the lowest observed Cohen's Kappa being $0.55$.

\paragraph{HateXplain} \cite{mathew_hatexplain_2021}. This dataset consists of posts from the social media platforms Twitter and Gab. Besides the message-level toxicity annotations, the annotators were also asked to ``highlight the rationales that could justify the final class.'' No inter-annotator agreement is reported for the span annotations.

\begin{table}[bthp]
    \centering
    \caption{Dataset statistics. Columns `Train', `Dev', and `Test' show the distribution of toxic (Toxic) and non-toxic ($\neg$Toxic) spans. The rows show the fraction of data that has spans (Span) and the fraction that does not (No span). The last column shows the average percentage of each sample's text that is part of a toxic span.}
    \let\mc\multicolumn
    \let\mr\multirow
    \setlength{\tabcolsep}{5pt}
    \begin{tabular}{@{}llrrrrrrr@{}}
        \toprule
   &         & \mc{1}{l}{Train} & \mc{1}{l}{}            & \mc{1}{l}{Dev}   & \mc{1}{l}{}            & \mc{1}{l}{Test}  & \mc{1}{l}{}            & \mr{2}{*}{Span-\%} \\ 
   \cmidrule(r){3-4} \cmidrule(r){5-6} \cmidrule(r){7-8} 
   &         & \mc{1}{l}{Toxic} & \mc{1}{l}{$\neg$Toxic} & \mc{1}{l}{Toxic} & \mc{1}{l}{$\neg$Toxic} & \mc{1}{l}{Toxic} & \mc{1}{l}{$\neg$Toxic} &\\ 
        \midrule
    \mr{2}{*}{SemEval}
               & Span    & 93.9\%       & -\,\,         & 93.8\%        & -\,\,         & 80.3\%       & -\,\,    & 13.2\% \\
               & No span & 6.1\%        & -\,\,         & 6.2\%         & -\,\,         & 19.7\%       & -\,\,    & -\,\, \\
        \cmidrule{2-9}
    \mr{2}{*}{HateXplain}
               & Span    & 57.6\%       & -\,\,         & 57.4\%        & -\,\,         & 57.5\%       & -\,\,    & 15.7\% \\
               & No span & 1.8\%        & 40.6\%        & 2.0\%         & 40.6\%        & 1.9\%        & 40.6\%  & -\,\, \\
        \bottomrule
    \end{tabular}
    \label{tab:dataset_balance}
\end{table}

\vspace{0.3\baselineskip}\noindent
In \autoref{tab:dataset_balance}, one can see that  both datasets have toxic samples annotated with toxic spans. 
However, the SemEval data does not include any non-toxic samples.
Furthermore, both datasets have some toxic samples without any spans (6.1\% and 1.8\%,  respectively). 
For both datasets this could either indicate that the annotators disagreed on which characters/tokens were toxic (final annotation was decided by a majority vote) or that the annotators agreed that, despite the sample being toxic, there is no specific span that is responsible for the toxicity of the message (implicit toxicity).

In order to perform the evaluation in the `ToxicInferred' setting, we train a binary toxicity classifier.
To make this possible on the SemEval dataset, we supplemented the data with non-toxic samples from the same Civil Comments data that the original dataset is based on.
In line with the requirements used for collecting the SemEval data, we take comments that were marked not toxic by a majority of at least three raters.
We randomly sample from the eligible comments until we reach a 50/50 balance between toxic and non-toxic messages.

\subsection{Implementation Details}
We use BERT-base \cite{devlin-etal-2019-bert} in the following three cases. 
After fine-tuning for binary toxicity classification we use it (1) as the model to which we apply rationale extraction and (2) for the binary toxicity predictions that are required for the `ToxicityInferred' setting. 
Finally, we also fine-tune BERT directly for toxic spans detection, including a variant with a final Conditional Random Fields (CRF) layer \cite{lafferty_conditional_2001}.
We choose BERT because Zhu et al. \cite{zhu-etal-2021-hitsz} used it to obtain state-of-the-art performance in the Semeval 2021 shared task \cite{pavlopoulos-etal-2021-semeval}.

\subsection{Hyper-parameter Search}
We first evaluate each combination of hyper-parameters using the same dataset for training and evaluation (in-domain). 
The training and evaluation are done on the canonical training and development splits, respectively.
To perform the hyper-parameter tuning, we select the set of hyper-parameter values with the best in-domain performance.
These are then used to evaluate on the test splits of both the same dataset (in-domain) and cross-domain dataset.

\subsubsection{Method-agnostic.}\label{sec:method-agnostic-hyper-parameter}
We include one hyper-parameter that influences the way in which the predicted spans are evaluated, determining how close together different spans are allowed to be. 
This process merges any two spans that are at most $n$ characters apart, which may be beneficial for each of the methods, since none of them predicts white space between tokens as toxic (the lexicons just match the words, while the other two methods use BERT tokenization which removes white space characters).
The grid-search values are $n \in \{ 0, 1, 9\,999\}$.
A value of $9\,999$ is added to join all spans together, never allowing more than one span.

\subsubsection{Lexicons.}\label{sec:lexicon-hyper-parameters}
We evaluate both constructed and existing lexicons.
The existing lexicons we use are HurtLex \cite{cabrio_hurtlex_2018} and the lexicon published by Wiegand et al. \cite{wiegand-etal-2018-inducing}.
Both lexicons come in two differently sized variants: `conservative' and `inclusive' for HurtLex, and `base' and `expanded' for Wiegand et al. \cite{wiegand-etal-2018-inducing}.
We refer to these as Hurtlex-c, Hurtlex-i, Wiegand-b, and Wiegand-e.
The constructed lexicons have method-specific hyper-parameters. 
The first is the threshold $\theta$ that sets the minimum toxicity score required for a word to enter the lexicon (see \autoref{sec:method:lexicon}). 
The second is the minimum number of occurrences of words in the dataset (min\_occ). 
We thereby exclude words that occur so infrequently that we cannot accurately measure their toxicity. 
Values included in the search are:
$\{0, 0.05, \dots, 1\}$ for the value of $\theta$, and $\{ 1, 3, 5, 7, 11 \}$ for the value of the minimum number of occurrences.

\subsubsection{Rationales.}\label{sec:rationale-extraction-hyper-parameters}
We include the following four input attribution methods in our experiments: Saliency \cite{shrikumar_learning_2017}, Integrated Gradients \cite{sundararajan_axiomatic_2017}, DeepLIFT \cite{simonyan_deep_2014}, and LIME \cite{ribeiro-etal-2016-trust}.
Each method works by generating scores that indicate the relative importance of the input tokens. 
Following Pluci\'{n}ski \& Klimczak \cite{plucinski-klimczak-2021-ghost}, we rescale the scores to sum up to 1.
The threshold that the score must exceed in order to be predicted as toxic is a hyper-parameter that we tune. 
Values included in the search for the threshold are $\{-0.05, -0.025, \dots, 0.5\}$.

\subsubsection{LMs.}
The hyper-parameters specific to the language models such as learning rate, dropout, etc. are left to their default values\footnote{See \url{https://huggingface.co/bert-base-cased/blob/main/config.json}}.

\begin{table}[p]
\setlength{\tabcolsep}{2.25pt}

\caption{Results for setting `ToxicOracle' after hyper-parameter tuning for $F_1^+$ on the Toxic part of each dataset. 
The metric columns from left to right are: $F_1^+$, Precision, and Recall on the Toxic part of the datasets; the $F_1^+$ score on the non-toxic part of the datasets ($\neg$Toxic); and, the macro average (harmonic mean) of the $F_1^+$ scores between the toxic and non-toxic parts of the dataset. 
The last two of which are in gray to emphasize that in this setting these metrics are not optimized and/or tuned for.
For both tables the overall highest scores are in bold, the best scores of the second best method are underlined.}

\begin{subtable}[h]{\textwidth}\centering
\caption{In-domain results for the SemEval and HateXplain datasets.}

\begin{tabu}{@{}l@{\hspace{0.2cm}}lrrr>{\color{gray}}r>{\color{gray}}rp{0.3cm}rrr>{\color{gray}}r>{\color{gray}}r@{}}
\toprule
&             & \multicolumn{3}{l}{Toxic}                     & $\neg$Toxic & Macro                       && \multicolumn{3}{l}{Toxic}                     & $\neg$Toxic   & Macro       \\ 
\cmidrule(r){3-5} \cmidrule(r){6-6} \cmidrule(){7-7} \cmidrule(r){9-11} \cmidrule(r){12-12} \cmidrule(){13-13} 
&             & $F_1^+$\,\,   & Prec.         & Rec.          & $F_1^+$\,\,   & $F_1^+$\,\,               && $F_1^+$\,\,   & Prec.         & Rec.          & $F_1^+$\,\,   & $F_1^+$\,\, \\ 
\specialrule{.5pt}{2pt}{2.5mm}
\midrule
&             & \multicolumn{5}{c}{HateXplain}                                                             && \multicolumn{5}{c}{SemEval}                                                \\ 
\midrule
\multirow{5}{*}{\rotatebox[origin=c]{90}{Lexicons}}
& Constr.    & {\ul 64.7}    & {\ul 74.4}    & 69.6          & 12.4           & 20.8                       && {\ul 59.8}    & 59.5         & {\ul 84.5}    & 59.0           & 59.4       \\
& HurtLex-c  & 36.4          & 47.2          & 39.5          & 14.2           & 20.4                       && 42.9          & 40.4         & 72.2          & 20.0           & 27.3       \\
& HurtLex-i  & 40.3          & 39.5          & 56.5          & 2.7            & 5.0                        && 34.9          & 29.0         & 76.6          & 6.3            & 10.6       \\
& Wiegand-b  & 47.4          & 68.2          & 48.4          & 37.5           & 41.8                       && 36.1          & 44.7         & 42.1          & 58.3           & 44.6       \\
& Wiegand-e  & 44.9          & 56.3          & 50.4          & 15.7           & 23.3                       && 46.1          & 44.2         & 73.7          & 21.9           & 29.7       \\
\midrule\multirow{4}{*}{\rotatebox[origin=c]{90}{Rationales}}
& Saliency   & 44.1          & 50.1          & 61.7          & 5.5            & 9.8                        && 53.6          & 57.5         & 73.4          & 26.8           & 35.7       \\
& Int. Grad. & 54.0          & 69.9          & 57.1          & 7.0            & 12.4                       && 57.6          & {\ul 60.2}   & 77.0          & 20.4           & 30.1       \\
& DeepLIFT   & 27.3          & 22.7          & {\ul 70.0}    & 4.9            & 8.2                        && 33.7          & 33.4         & 54.3          & 9.2            & 14.5       \\
& LIME       & 40.6          & 46.6          & 46.1          & 0.3            & 0.5                        && 45.9          & 48.1         & 63.1          & 52.0           & 48.7       \\
\midrule\multirow{2}{*}{\rotatebox[origin=c]{90}{LMs}}
& BERT       & \textbf{74.9} & \textbf{82.3} & \textbf{80.4} & 12.3           & 21.1                       && \textbf{64.4} & \textbf{64.7}& \textbf{87.4} & 51.0           & 56.9       \\
& BERT+CRF   & 73.5          & 80.8          & 79.3          & 12.1           & 20.9                       && 64.1          & 64.5         & 86.7          & 50.4           & 56.4       \\
\bottomrule
\end{tabu}
\label{tab:in-domain-toxic-results}
\end{subtable}

\begin{subtable}[h]{\textwidth}\centering
\caption{Cross-domain results for the SemEval and HateXplain datasets. Column title $X \rightarrow Y$ indicates trained on $X$, evaluated on $Y$. }

\begin{tabu}{@{}l@{\hspace{0.2cm}}lrrr>{\color{gray}}r>{\color{gray}}rp{0.3cm}rrr>{\color{gray}}r>{\color{gray}}r@{}}
\midrule
&             & \multicolumn{5}{c}{SemEval $\rightarrow$ HateXplain}                                       && \multicolumn{5}{c}{HateXplain $\rightarrow$ SemEval}                           \\
\midrule
\rowfont{\color{white}}
&             & \multicolumn{3}{l}{Toxic}                     & $\neg$Toxic & Macro                        && \multicolumn{3}{l}{Toxic}                     & $\neg$Toxic   & Macro          \\ 
\rowfont{\color{white}}
&             & $F_1^+$\,\,   & Prec.         & Rec.          & $F_1^+$\,\,   & $F_1^+$\,\,                && $F_1^+$\,\,   & Prec.         & Rec.          & $F_1^+$\,\,   & $F_1^+$\,\,    \\[-0.8cm] 
\multirow{5}{*}{\rotatebox[origin=c]{90}{Lexicons}}
& Constr.     & 24.6          & 49.7          & 23.1          & 45.0          & 31.8                       && 13.6          & 16.7          & 9.1           & 46.5          & 21.0           \\
& HurtLex-c   & 36.4          & 47.2          & 39.5          & 14.2          & 20.4                       && 42.9          & 40.4          & 72.2          & 20.0          & 27.3           \\
& HurtLex-i   & 40.3          & 39.5          & \textbf{56.5} & 2.7           & 5.0                        && 34.9          & 29.0          & \textbf{76.6} & 6.3           & 10.6           \\
& Wiegand-b   & \textbf{47.4} & \textbf{68.2} & 48.4          & 37.5          & 41.8                       && 36.1          & \textbf{44.7} & 42.1          & 58.3          & 44.6           \\
& Wiegand-e   & 44.9          & 56.3          & 50.4          & 15.7          & 23.3                       && \textbf{46.1} & 44.2          & 73.7          & 21.9          & 29.7           \\
\midrule
\multirow{4}{*}{\rotatebox[origin=c]{90}{Rationales}}
& Saliency    & 39.0          & 53.8          & 41.2          & 6.8           & 11.5                       && 33.2          & 28.3          & {\ul 63.8}    & 28.7          & 30.8           \\
& Int. Grad.  & 34.2          & 44.0          & 37.6          & 2.7           & 5.0                        && {\ul 35.0}    & {\ul 33.1}    & 61.1          & 12.8          & 18.7           \\
& DeepLIFT    & 27.2          & 33.6          & 34.6          & 3.1           & 5.5                        && 17.5          & 13.9          & 63.5          & 11.7          & 14.0           \\
& LIME        & 23.5          & 34.4          & 24.8          & 8.2           & 12.1                       && 17.6          & 15.4          & 32.4          & 0.9           & 1.6            \\
\midrule
\multirow{2}{*}{\rotatebox[origin=c]{90}{LMs}}
& BERT        & 42.7          & 56.7          & 45.9          & 16.8          & 24.1                       && 25.7          & 31.5          & 31.8          & 60.5          & 36.0           \\
& BERT+CRF    & {\ul 42.8}    & {\ul 57.6}    & {\ul 46.1}    & 18.5          & 25.9                       && 27.5          & 29.3          & 40.6          & 27.7          & 27.6           \\
\bottomrule
\end{tabu}
\label{tab:cross-domain-toxic-results}
\end{subtable}

\end{table}
\begin{table}
\caption{Results for the `ToxicInferred' setting after hyper-parameter tuning for Macro $F_1^+$. The metric columns from left to right are: $F_1^+$, Precision, and Recall on the Toxic part of the datasets; the $F_1^+$ score on the non-toxic part of the datasets ($\neg$Toxic); and, the macro average (harmonic mean) of the $F_1^+$ scores between the toxic and non-toxic parts of the dataset. In both tables, the overall highest scores are in bold, the best scores of the second best method are underlined.}

\setlength{\tabcolsep}{2.2pt}

\begin{subtable}[h]{\textwidth}\centering
\caption{In-domain results for the SemEval and HateXplain datasets.}

\begin{tabu}{@{}l@{\hspace{0.2cm}}lrrrrrp{0.3cm}rrrrr@{}}
\toprule
&             & \multicolumn{3}{l}{Toxic}                     & $\neg$Toxic & \multicolumn{1}{c}{Macro}   && \multicolumn{3}{l}{Toxic}                     & $\neg$Toxic   & \multicolumn{1}{c}{Macro} \\ 
\cmidrule(r){3-5} \cmidrule(r){6-6} \cmidrule(){7-7} \cmidrule(r){9-11} \cmidrule(r){12-12} \cmidrule(){13-13} 
&             & $F_1^+$\,\,   & Prec.         & Rec.          & $F_1^+$\,\,   & $F_1^+$\,\,               && $F_1^+$\,\,   & Prec.         & Rec.          & $F_1^+$\,\,   & $F_1^+$\,\,               \\ 
\specialrule{.5pt}{2pt}{2.5mm}
\midrule
&             & \multicolumn{5}{c}{HateXplain}                                                             && \multicolumn{5}{c}{SemEval}                                                              \\ 
\midrule
\multirow{5}{*}{\rotatebox[origin=c]{90}{Lexicons}}
& Constr.     & {\ul 53.3}    & {\ul 81.1}    & 53.2          & 82.0          & {\ul 64.6}                &  & {\ul 60.5}    & {\ul 61.9}    & {\ul 81.2}    & 95.8          & {\ul 74.2}              \\
& HurtLex-c   & 30.8          & 48.4          & 31.8          & 82.6          & 44.8                      &  & 43.7          & 41.2          & 70.6          & 95.2          & 59.9                    \\
& HurtLex-i   & 33.8          & 40.9          & 45.7          & 81.5          & 47.7                      &  & 35.7          & 29.4          & 74.2          & 94.8          & 51.9                    \\
& Wiegand-b   & 42.9          & 72.3          & 42.9          & \textbf{84.5} & 56.9                      &  & 37.0          & 45.9          & 41.4          & \textbf{97.0} & 53.6                    \\
& Wiegand-e   & 41.2          & 61.1          & 44.7          & 82.2          & 54.9                      &  & 46.7          & 45.1          & 71.9          & 95.3          & 62.7                    \\ 
\midrule
\multirow{4}{*}{\rotatebox[origin=c]{90}{Rationales}}
& Saliency    & 36.0          & 53.3          & 46.8          & {\ul 82.7}    & 50.1                      &  & 53.5          & 58.4          & 70.4          & 94.7          & 68.4                    \\
& Int. Grad.  & 46.7          & 80.3          & 45.1          & 81.3          & 59.4                      &  & 57.9          & 61.7          & 74.4          & 94.7          & 71.9                    \\
& DeepLIFT    & 21.6          & 21.5          & {\ul 54.8}    & 81.7          & 34.1                      &  & 34.4          & 33.8          & 52.2          & 94.8          & 50.5                    \\
& LIME        & 37.4          & 59.1          & 37.5          & 81.3          & 51.2                      &  & 47.1          & 49.8          & 61.8          & 94.6          & 62.8                    \\ 
\midrule
\multirow{2}{*}{\rotatebox[origin=c]{90}{LMs}}
& BERT        & \textbf{59.5} & \textbf{84.9} & \textbf{61.7} & 81.5          & \textbf{68.7}             &  & \textbf{63.9} & \textbf{65.6} & \textbf{84.1} & 94.8          & \textbf{76.4}           \\
& BERT+CRF    & 58.6          & 83.6          & 61.1          & 81.7          & 68.3                      &  & 63.6          & 65.4          & 83.3          & {\ul 94.9}    & 76.1                    \\
\bottomrule
\end{tabu}
\label{tab:in-domain-macro-results}
\end{subtable}

\begin{subtable}[h]{\textwidth}\centering
\caption{Cross-domain results for the SemEval and HateXplain datasets. Column title $X \rightarrow Y$ indicates trained on $X$, evaluated on $Y$. }
\begin{tabu}{@{}l@{\hspace{0.2cm}}lrrrrrp{0.3cm}rrrrr@{}}
\midrule
&             & \multicolumn{5}{c}{SemEval $\rightarrow$ HateXplain}                                      && \multicolumn{5}{c}{HateXplain $\rightarrow$ SemEval}                                      \\
\midrule
\rowfont{\color{white}}
&             & \multicolumn{3}{l}{Toxic}                     & $\neg$Toxic & \multicolumn{1}{c}{Macro}   && \multicolumn{3}{l}{Toxic}                     & $\neg$Toxic   & \multicolumn{1}{c}{Macro} \\ 
\rowfont{\color{white}}
&             & $F_1^+$\,\,   & Prec.         & Rec.          & $F_1^+$\,\,   & $F_1^+$\,\,               && $F_1^+$\,\,   & Prec.         & Rec.          & $F_1^+$\,\,   & $F_1^+$\,\,               \\[-0.8cm] 
\multirow{5}{*}{\rotatebox[origin=c]{90}{Lexicons}}
& Constr.     & 14.3          & 43.6          & 11.1          & \textbf{64.2} & 23.4                      && 17.2          & 18.2          & 2.6           & 97.7          & 29.3                      \\
& HurtLex-c   & 29.5          & 51.2          & 30.8          & 51.9          & 37.6                      && 23.1          & 34.2          & 20.0          & 96.4          & 37.3                      \\
& HurtLex-i   & 30.6          & 41.8          & \textbf{40.6} & 48.6          & 37.5                      && 21.3          & 25.3          & \textbf{21.6} & 96.1          & 34.9                      \\
& Wiegand-b   & \textbf{34.5} & \textbf{66.6} & 34.4          & 62.0          & \textbf{44.3}             && 22.7          & \textbf{39.2} & 13.2          & \textbf{97.8} & 36.8                      \\
& Wiegand-e   & 31.8          & 56.0          & 34.3          & 54.2          & 40.1                      && \textbf{23.4} & 35.2          & 19.9          & 96.4          & \textbf{37.6}             \\ 
\midrule
\multirow{4}{*}{\rotatebox[origin=c]{90}{Rationales}}
& Saliency    & 27.3          & 53.6          & 27.0          & 49.5          & 35.2                       && 21.1          & 28.2          & 15.8          & {\ul 96.4}    & 34.6                     \\
& Int. Grad.  & 25.4          & 47.7          & 24.4          & 48.3          & 33.3                       && {\ul 22.4}    & {\ul 34.7}    & 15.3          & 96.0          & {\ul 36.3}               \\
& DeepLIFT    & 19.7          & 34.3          & 22.7          & 49.0          & 28.1                       && 16.5          & 11.5          & {\ul 19.3}    & 96.0          & 28.1                     \\
& LIME        & 20.1          & 40.7          & 19.2          & 49.0          & 28.5                       && 19.8          & 23.9          & 13.6          & 96.0          & 32.9                     \\ 
\midrule
\multirow{2}{*}{\rotatebox[origin=c]{90}{LMs}}
& BERT        & 31.6          & 55.7          & 33.3          & 51.3          & 39.1                      && 20.3          & 27.0          & 11.8          & {\ul 96.4}    & 33.5                      \\
& BERT+CRF    & {\ul 32.1}    & {\ul 56.7}    & {\ul 33.8}    & {\ul 51.7}    & {\ul 39.6}                && 20.1          & 25.9          & 13.1          & 96.2          & 33.2                      \\
\bottomrule
\end{tabu}
\label{tab:cross-domain-macro-results}
\end{subtable}
\end{table}

\section{Results}
In this section, we present the results of our experiments.
We first report the in-domain performance of the span detection methods. 
Then we report the cross-domain performance and the relative drop compared to the in-domain results.

\subsubsection{In-domain.}\label{sec:in_domain_macr_results}
Performance of the methods can be seen in \autoref{tab:in-domain-toxic-results} for the `ToxicOracle' setting, and in \autoref{tab:in-domain-macro-results} for the `ToxicInferred' setting.
We observe similar patterns in both settings.
For example, it is clear that in both cases in-domain performance is highest for the fine-tuned LMs, which matches results obtained in the shared task \cite{pavlopoulos-etal-2021-semeval}.
The second best scores are achieved with the lexicons constructed from span-annotated training data.
Existing lexicons do worse and are outperformed by the rationale extraction using Integrated Gradients.

When comparing our results (\autoref{tab:in-domain-toxic-results}) to those obtained by Zhu et al. \cite{zhu-etal-2021-hitsz}, we see that our fine-tuned LMs and lexicon underperform theirs by several points (64.4 vs. 69.44 for the LMs and 59.8 vs. 65.0 for the lexicon). This could be because we did not clean the training data as they did or due to minor differences in training setup and lexicon construction.

\subsubsection{Cross-domain.}\label{sec:cross_domain_results} 
The performance of the methods under cross-domain conditions can be seen in \autoref{tab:cross-domain-toxic-results} for the `ToxicOracle' setting and in \autoref{tab:cross-domain-macro-results} for the `ToxicInferred' setting.
Contrary to the in-domain results, the fine-tuned LMs are outperformed by the Wiegand et al. \cite{wiegand-etal-2018-inducing} lexicons in all cases.

We calculate the ratio of cross-domain performance to in-domain performance (as measured by Toxic and Macro $F_1^+$ scores for the `ToxicOracle' and `ToxicInferred' settings, respectively).
The performance of the constructed lexicons drops dramatically (to 34\% of the in-domain scores on average) resulting in them being ranked last in the cross-domain setup.
The LMs retain more of their performance, but still drop to (on average) 50\%. The rationale extraction methods keep 62\% of their original performance on average.
Since the existing lexicons are not related to any domain, they do not lose any performance in the `ToxicOracle' setting.
In the `ToxicInferred' setting the drop is small for `SemEval $\rightarrow$ HateXplain' (retaining 86\%) while losing substantial performance for `HateXplain $\rightarrow$ SemEval' (keeping only 56\%).
The only reason these lexicons could perform worse in this setting is due to cross-domain application of the binary toxicity classifier, suggesting that the classifier transfers much better in one direction than the other.

\section{Error Analysis}\label{sec:error_analysis}
We analyse and compare the types of errors made by each of the methods. 
We take inspiration from van Aken et al. \cite{van-aken-etal-2018-challenges} who perform a detailed error analysis where they classify errors by their type.
We analyse prediction errors made by the best performing variant of each method. 
By selecting the best methods we analyse the best case scenario for each approach.
The errors are sampled such that we have guaranteed representation for every combination of high and low precision and recall%
\ifappendices (see \autoref{apx:sec:error_analysis_details} for details)\fi.
We sample 75 errors for each method on each dataset (225 per dataset, 450 total).
We identify a number of error classes, where each contains either false negatives (FN) or false positives (FP).
Four classes and three aggregations can be seen with their prevalence for each method and dataset in \autoref{tab:error_analysis_results}.

\subsubsection{Doubtful Labels.}
Likely due to the subjective nature of this task, the number of errors classified as having a doubtful label was quite high. 
In total, 40.9\% of the sampled HateXplain errors, and 23.5\% of the sampled SemEval  errors had a doubtful label.
This is in line with analyses done for message-level detection \cite{markov_ensemble_2022}.

\subsubsection{False Negatives.}
The language model has the lowest false negative rate for HateXplain, but for the SemEval dataset the lexicon-based span prediction has the lowest false negative rate. 
The \texttt{FN-explicit} class indicates what proportion of false negatives involved explicitly toxic words (e.g.,  ``nonsensical aussie \textbf{retarded} babbles'').
The class was applied to any prediction that involved not predicting a word despite it being explicitly toxic.
On both datasets, these kinds of errors were most common for the rationale extraction method, and least common for the lexicon-based predictions. 
The latter was expected since these lexicons are created specifically to cover explicitly toxic words.
We also tracked what we call subword errors, which are span predictions that do not cover a word entirely. 
The \texttt{FN-subword-toxic} class was applied to any erroneous spans from which a morphologically relevant part was missing.
For example: ``\dots what \textbf{stupid}ity and arrogance \dots'' (predicted span in bold).
These errors were most prevalent among the lexicon-based predictions. 
This is due to the lexicons being applied by finding exact matches without taking into account affixes.
\subsubsection{False Positives.}
The overall false positive rate is the lowest for the rationale extraction method on both datasets, and was high for the lexicons and LMs. 
A high false positive rate for LMs is in line with previous findings on message-level toxicity detection \cite{markov-daelemans-2021-improving}. 
For the lexicon the high rate can be explained by the high rate of \texttt{FP-subword-toxic} errors.
That class tracks false positives where one of the spans is an explicitly toxic word, but inside a non-toxic word, for example, the words `ho' and `lame' being marked in: ``\dots that I some\textbf{ho}w b\textbf{lame} him \dots''.
This happens often for the lexicon predictions, because it looks for any matches with the lexicon's entries.
We also included \texttt{FP-target}, which is a false positive of a target group, for example: ``\dots \textbf{republican} you are not welcome here \dots''.
This error type is quite rare, but more common for the fine-tuned LMs.

\begin{table}[bthp]
\centering
\let\mc\multicolumn
\let\mr\multirow
\setlength{\tabcolsep}{1pt}

\caption{The results of the error analysis, showing the prevalence of each class (rows) for every method on each dataset (columns). 
Last three rows show aggregates, with percentage of errors that had any of the subword classes, false negative classes, or false positive classes.
\ifappendices List of classes included in the aggregates can be found in \autoref{apx:sec:error_analysis_details}.\fi%
}

\begin{tabular}{@{}lrrrrrrr@{}}
\toprule
   & \mc{3}{c}{SemEval $\rightarrow$ HateXplain} &  & \mc{3}{c}{HateXplain $\rightarrow$ SemEval} \\ 
\cmidrule{2-4}\cmidrule{6-8}
    & \mc{1}{c}{Lexicon}& \mc{1}{c}{Rationale} & \mc{1}{c}{LM} &  & \mc{1}{c}{Lexicon} & \mc{1}{c}{Rationale} & \mc{1}{c}{LM} \\ 
    & \mc{1}{c}{(Wiegand-b)}& \mc{1}{c}{(Saliency)} & \mc{1}{c}{(BERT)} &  & \mc{1}{c}{(Wiegand-e)} & \mc{1}{c}{(Int. Grad.)} & \mc{1}{c}{(BERT)} \\ 
\midrule
\texttt{FN-explicit}             & 19.6\%         & 36.5\%           & 21.4\%           &  & 1.7\%          & 37.3\%           & 32.3\%   \\
\texttt{FN-subword-morph}        & 34.5\%         & 4.9\%            & 0.0\%            &  & 11.9\%         & 4.0\%            & 0.0\%    \\
\texttt{FP-subword-toxic}        & 5.5\%          & 2.1\%            & 0.8\%            &  & 31.1\%         & 0.9\%            & 3.6\%    \\
\texttt{FP-target}               & 0.0\%          & 3.1\%            & 5.5\%            &  & 0.7\%          & 0.6\%            & 2.2\%    \\
\midrule
\texttt{*-subword-*}             & 16.7\%         & 16.3\%           & 2.5\%            &  & 66.5\%         & 10.2\%           & 12.0\%   \\ 
\texttt{FN-*}                    & 59.9\%         & 56.5\%           & 48.1\%           &  & 23.6\%         & 60.4\%           & 52.6\%   \\
\texttt{FP-*}                    & 32.5\%         & 22.3\%           & 48.2\%           &  & 76.8\%         & 25.9\%           & 60.9\%   \\
\bottomrule
\end{tabular}
\label{tab:error_analysis_results}
\end{table}

\section{Conclusion \& Discussion}
We have evaluated three kinds of methods for toxic spans predictions in a cross-domain setting. 
Our results show that the performance of the fine-tuned LMs suffers greatly when applied to out-of-domain data, thereby making off-the-shelf lexicons of toxic language the best performing option. 
This suggests that fine-tuned LMs do not handle domain shift that may occur from changes in the use of toxic language or the relative prominence of communities in the data.
This differs from what was observed for the message-level task, where LMs showed better generalization capabilities.
The cross-domain error analysis showed that language models are more likely to produce false positives (excluding subword false positives). This means that tokens that are toxic in the training data are not toxic in the test data across domains, where the learned lexical representations do not transfer and are also not corrected in context by the models. In some cases, we also found that targets of toxic language were falsely included in the predicted spans.
On the other hand, the spans predicted by language models also miss more explicit toxicity than those predicted with lexicons, although rationale extraction misses even more still.

Limitations of this work include: (a) the fine-tuning approach being evaluated with BERT and no other LM; (b) the absence of attention-based XAI methods among those selected for the rationale extraction approach; and (c) having no more than two span-annotated datasets for the cross-domain evaluation.

In future work, we will focus on improving cross-domain performance by combining approaches explored in this work within an ensemble strategy, since our error analysis suggests that the methods make different types of errors.

\subsubsection{Acknowledgements.}
This research was supported by Huawei Finland through the DreamsLab project. All content represented the opinions of the authors, which were not necessarily shared or endorsed by  their respective employers and/or sponsors.

\bibliography{references,anthology-abbr}
\bibliographystyle{splncs04}

\FloatBarrier

\ifappendices \newpage

\appendix

\renewcommand{\thetable}{\Alph{section}.\arabic{table}}

\setcounter{table}{0}
\section{Additional result table}
\begin{table}[!b]
\centering
\caption{Setting `ToxicOracle': results after hyper-parameter tuning for macro $F_1^+$. The metric columns from left to right are: $F_1^+$, precision, and recall on the toxic part of the datasets; the $F_1^+$ score on the non-toxic part of the datasets ($\neg$Toxic); and, the macro average (harmonic mean) of the $F_1^+$ scores between the toxic and non-toxic parts of the dataset. In both tables, the overall highest scores are in bold, the best scores of the second best method are underlined.}

\setlength{\tabcolsep}{2.2pt}

\begin{subtable}[h]{\textwidth}\centering
\caption{In-domain results for the SemEval and HateXplain datasets.}
\begin{tabu}{@{}l@{\hspace{0.2cm}}lrrrrrp{0.3cm}rrrrr@{}}
\toprule
&             & \multicolumn{3}{l}{Toxic}                     & $\neg$Toxic & \multicolumn{1}{c}{Macro}   && \multicolumn{3}{l}{Toxic}                     & $\neg$Toxic   & \multicolumn{1}{c}{Macro} \\ 
\cmidrule(r){3-5} \cmidrule(r){6-6} \cmidrule(){7-7} \cmidrule(r){9-11} \cmidrule(r){12-12} \cmidrule(){13-13} 
&             & $F_1^+$\,\,   & Prec.         & Rec.          & $F_1^+$\,\,   & $F_1^+$\,\,               && $F_1^+$\,\,   & Prec.         & Rec.          & $F_1^+$\,\,   & $F_1^+$\,\,               \\ 
\specialrule{.5pt}{2pt}{2.5mm}
\midrule
&             & \multicolumn{5}{c}{HateXplain}                                                             && \multicolumn{5}{c}{SemEval}                                                              \\ 
\midrule
\multirow{5}{*}{\rotatebox[origin=c]{90}{Lexicons}}
& Constr.     & {\ul 54.2}    & \textbf{93.2} & 49.1          & \textbf{73.7} & \textbf{62.5}              && {\ul 63.1}   & \textbf{69.0} & {\ul 77.5}    & \textbf{83.5} & \textbf{71.9}             \\
& HurtLex-c   & 36.4          & 47.2          & 39.5          & 14.2          & 20.4                       && 42.9         & 40.4          & 72.2          & 20.0          & 27.3                      \\
& HurtLex-i   & 40.3          & 39.5          & 56.5          & 2.7           & 5.0                        && 34.9         & 29.0          & 76.6          & 6.3           & 10.6                      \\
& Wiegand-b   & 47.4          & 68.2          & 48.4          & 37.5          & 41.8                       && 36.1         & 44.7          & 42.1          & 58.3          & 44.6                      \\
& Wiegand-e   & 44.9          & 56.3          & {\ul 50.4}    & 15.7          & 23.3                       && 46.1         & 44.2          & 73.7          & 21.9          & 29.7                      \\
\midrule
\multirow{4}{*}{\rotatebox[origin=c]{90}{Rationales}}
& Saliency    & 30.1          & 70.9          & 28.3          & 44.6          & 35.9                       && 48.7         & {\ul 67.9}    & 52.5          & 52.8          & 50.7                      \\
& Int. Grad.  & 34.3          & 83.9          & 27.9          & 40.2          & {\ul 37.0}                 && 53.2         & 74.3          & 53.4          & {\ul 70.7}    & 60.7                      \\
& DeepLIFT    & 21.1          & 33.2          & 39.3          & {\ul 48.8}    & 29.4                       && 33.3         & 42.6          & 34.0          & 45.1          & 38.3                      \\
& LIME        & 30.7          & 62.2          & 34.2          & 36.3          & 33.3                       && 47.7         & 58.1          & 53.9          & 67.9          & 56.0                      \\
\midrule
\multirow{2}{*}{\rotatebox[origin=c]{90}{LMs}}
& BERT       & \textbf{74.9} & {\ul 82.3}    & \textbf{80.4} & 12.3           & 21.1                       && \textbf{64.4}& 64.7          & \textbf{87.4} & 50.9          & {\ul 56.9}                \\
& BERT+CRF   & 73.5          & 80.8          & 79.3          & 12.1           & 20.9                       && 64.1         & 64.5          & 86.7          & 50.4          & 56.4                      \\
\bottomrule
\end{tabu}
\label{tab:in-domain-macro-noprop-results}
\end{subtable}

\begin{subtable}[h]{\textwidth}\centering
\caption{Cross-domain results for the SemEval and HateXplain datasets. Column title $X \rightarrow Y$ indicates trained on $X$, evaluated on $Y$.}
\begin{tabu}{@{}l@{\hspace{0.2cm}}lrrrrrp{0.3cm}rrrrr@{}}
\midrule
&             & \multicolumn{5}{c}{SemEval $\rightarrow$ HateXplain}                                      && \multicolumn{5}{c}{HateXplain $\rightarrow$ SemEval}                                      \\
\midrule
\rowfont{\color{white}}
&             & \multicolumn{3}{l}{Toxic}                     & $\neg$Toxic & \multicolumn{1}{c}{Macro}   && \multicolumn{3}{l}{Toxic}                     & $\neg$Toxic   & \multicolumn{1}{c}{Macro} \\ 
\rowfont{\color{white}}
&             & $F_1^+$\,\,   & Prec.         & Rec.          & $F_1^+$\,\,   & $F_1^+$\,\,               && $F_1^+$\,\,   & Prec.         & Rec.          & $F_1^+$\,\,   & $F_1^+$\,\,               \\[-0.8cm] 
\multirow{5}{*}{\rotatebox[origin=c]{90}{Lexicons}}
& Constr.     & 12.8          & 46.6          & 9.4           & \textbf{65.1} & 21.4                      && 19.1          & 11.6          & 0.3           & \textbf{95.7} & 31.9                      \\
& HurtLex-c   & 36.4          & 47.2          & 39.5          & 14.2          & 20.4                      && 42.9          & 40.4          & 72.2          & 20.0          & 27.3                      \\
& HurtLex-i   & 40.3          & 39.5          & \textbf{56.5} & 2.7           & 5.0                       && 34.9          & 29.0          & \textbf{76.6} & 6.3           & 10.6                      \\
& Wiegand-b   & \textbf{47.4} & \textbf{68.2} & 48.4          & 37.5          & \textbf{41.8}             && 36.1          & {\ul 44.7}    & 42.1          & 58.3          & {\ul 44.6}                \\
& Wiegand-e   & 44.9          & 56.3          & 50.4          & 15.7          & 23.3                      && \textbf{46.1} & 44.2          & 73.7          & 21.9          & 29.7                      \\
\midrule
\multirow{4}{*}{\rotatebox[origin=c]{90}{Rationales}}
& Saliency   & 29.2          & {\ul 61.1}    & 24.3          & 30.6           & {\ul 29.9}                && {\ul 34.9}    & 47.2          & 34.5          & {\ul 65.9}    & \textbf{45.7}             \\
& Int. Grad. & 16.6          & 48.8          & 13.3          & {\ul 53.7}     & 25.4                      && 28.4          & \textbf{47.7} & 21.9          & 57.6          & 38.0                      \\
& DeepLIFT   & 20.0          & 41.9          & 19.8          & 28.5           & 23.5                      && 17.8          & 17.0          & 26.0          & 56.4          & 27.1                      \\
& LIME       & 17.6          & 37.4          & 16.5          & 27.5           & 21.5                      && 19.9          & 19.8          & 33.7          & 43.7          & 27.3                      \\ 
\midrule
\multirow{2}{*}{\rotatebox[origin=c]{90}{LMs}}
& BERT       & 42.7          & 56.7          & 45.9          & 16.8           & 24.1                      && 25.7          & 31.5          & 31.8          & 60.5          & 36.0                      \\
& BERT+CRF   & {\ul 42.8}    & 57.6          & {\ul 46.1}    & 18.5           & 25.9                      && 27.5          & 29.3          & {\ul 40.6}    & 27.7          & 27.6                      \\
\bottomrule
\end{tabu}
\label{tab:cross-domain-macro-noprop-results}
\end{subtable}

\end{table}

\FloatBarrier

\setcounter{table}{0}
\section{Error analysis details}\label{apx:sec:error_analysis_details}
The erroneous predictions are sampled as follows.
\begin{enumerate}[-,nosep]
    \item We start with the cross-domain erroneous predictions ($F_1 < 1$) of the best performing lexicon, attribution method, and language model on both the HateXplain and SemEval datasets (in the `ToxicOracle' setting). 
    \item We perform a non-uniform sampling where we categorize the predictions based on their precision and recall first, and then sample from each category to ensure that they are all represented.
    Splitting into low/high was done based on ranking, i.e., values up to and including the median are low, values above the median are high, yielding four categories. The relative frequency of these categories is corrected before error class prevalence is reported for the whole dataset.
    We include a final category for empty predictions, since precision and recall are not defined for these predictions. 
    \item For each category, we sampled 15 data points. Only one category (low precision and high recall) for one of the methods (fine-tuned LM trained on HateXplain, tested on SemEval) was empty, all others had at least 15 samples (also see \autoref{tab:error_analysis_details}). More samples were drawn from the other categories to compensate for this. 
\end{enumerate}
\vspace{1em}
We included the following error classes in our analysis:
\begin{itemize}[topsep=2pt]
    \item Doubtful labels:
    \begin{itemize}[topsep=0pt]
        \item \texttt{doubt-label-missing}: If the label spans do not include toxic spans that they should.
        \item \texttt{doubt-label-toomany}: If the label spans include spans that should not be included, because we do not think they are toxic.
    \end{itemize}

    \item Subword errors, indicating lack of word-level understanding:
    \begin{itemize}   
        \item \texttt{FP-subword-toxic}: incorrectly recognizes toxic word within non-toxic words. 
        For example: the words `ho' and `lame' being marked in: ``\dots the straw man argument that I some\textbf{ho}w b\textbf{lame} him \dots''.
        \item \texttt{FP-subword-nontoxic}: incorrectly recognizes a non-toxic word within other non-toxic words. 
        For example: ``\dots destroy records and burn the p\textbf{ape}rwork? \dots''
        \item \texttt{FN-subword-morph}: morphologically relevant parts of toxic span are not predicted, such as predicting only the stem of a toxic word. 
        For example: ``\dots what stupidity and arrogance \dots''.
        \item \texttt{FN-subword}: missing part of toxic word that is not morphologically relevant. 
        For example: only `oron' being predicted in `m\textbf{oron}''.
    \end{itemize}
    
    \item False negatives:
    \begin{itemize}
        \item \texttt{FN-explicit}: missing explicitly toxic spans; including slurs, etc.
        \item \texttt{FN-explicit-spelling}: missing explicitly toxic spans because of abbreviations, or uncommon or alternative spellings.
        \item \texttt{FN-implicit}: missing non-obvious toxic spans. 
        For example: ``<user> <user> i can’t stand this look they all look like identical blow up dolls''. Other examples include metaphors and irony.
        \item \texttt{FN-phrase-part}: missed part of (some of the words) in a toxic phrase. 
        For example: ``...  Posting his citation (on-line) only shows that "diesel \textbf{jerk}" is proud of his actions, rather than ashamed.'' does not include `diesel' in the toxic span.
        \item \texttt{FN-whitespace}: prediction does not include white space that it should have included.
    \end{itemize}
    
    \item False positives: 
    \begin{itemize}
        \item \texttt{FP-target}: predicting target groups (or proper nouns) as part of the toxic spans. 
        For example: ``\dots republican you are not welcome here we hate you \dots''.
        \item \texttt{FP-pos}: predicting words that are a part-of-speech which should not be part of toxic spans, such as pronouns and prepositions.
    \end{itemize}
\end{itemize}

\begin{table}[hb]
\setlength{\tabcolsep}{3.5pt}

\caption{The number of samples in each category, from left to right: high precision and recall, high precision and low recall, low precision and high recall, low precision and recall, empty predictions, and total number of errors.}

\begin{tabular}{@{}lrrrrrrrrrrrrr@{}}
\toprule
          & \multicolumn{5}{c}{SemEval $\rightarrow$ HateXplain}                                       &                        &  & \multicolumn{5}{c}{HateXplain $\rightarrow$ SemEval}                                       &                        \\ \cmidrule(lr){2-7} \cmidrule(l){9-14} 
          & $P\uparrow$ & $P\uparrow$   & $P\downarrow$ & $P\downarrow$ & \multirow{2}{*}{$\emptyset$} & \multirow{2}{*}{Total} &  & $P\uparrow$ & $P\uparrow$   & $P\downarrow$ & $P\downarrow$ & \multirow{2}{*}{$\emptyset$} & \multirow{2}{*}{Total} \\
          & $R\uparrow$ & $R\downarrow$ & $R\uparrow$   & $R\downarrow$ &                              &                        &  & $R\uparrow$ & $R\downarrow$ & $R\uparrow$   & $R\downarrow$ &                              &                        \\
\midrule
Lexicon   & 215         & 194           & 170           & 240           & 176                          & 995                    &  & 436         & 367           & 189           & 631           & 94                           & 1717                   \\
Rationale & 227         & 113           & 122           & 257           & 372                          & 1091                   &  & 158         & 92            & 17            & 234           & 1209                         & 1710                   \\
LM        & 263         & 122           & 112           & 282           & 152                          & 931                    &  & 545         & 39            & 0             & 854           & 348                          & 1786                   \\
\midrule
Lexicon   & 22\%        & 19\%          & 17\%          & 24\%          & 18\%                         &                  &  & 25\%        & 21\%          & 11\%          & 37\%          & 5\%                          &                  \\
Rationale & 21\%        & 10\%          & 11\%          & 24\%          & 34\%                         &                   &  & 9\%         & 5\%           & 1\%           & 14\%          & 71\%                         &                   \\
LM        & 28\%        & 13\%          & 12\%          & 30\%          & 16\%                         &                   &  & 31\%        & 2\%           & 0\%           & 48\%          & 19\%                         &                   \\
\bottomrule
\end{tabular}

\label{tab:error_analysis_details}
\end{table}

\FloatBarrier

\newpage

\setcounter{table}{0}
\section{Hyper-parameters}\label{apx:sec:hyperparameters}

\begin{table}[h!]

\caption{Best sets of hyperparameters when fine-tuning for Macro $F_1^+$ for each model on both datasets. `Fill-chars' is a hyper-parameter described in \autoref{sec:method-agnostic-hyper-parameter}.}

\begin{subtable}[htbp]{\textwidth}\centering

\caption{Lexicons. The method-specific parameters are described in \autoref{sec:lexicon-hyper-parameters}, with `Min-occ' referring to the minimum occurrences required for inclusion in the lexicon. }

\setlength{\tabcolsep}{4.5pt}
\begin{tabular}{@{}llrrrrrrr@{}}
\toprule
\multirow{2}{*}{Prop-binary}
    &           &  \multicolumn{3}{l}{HateXplain}  & & \multicolumn{3}{l}{SemEval}  \\ 
\cmidrule(r){3-5} \cmidrule(){7-9} 
    &           & Fill-chars & Min-occ. & $\theta$  &  & Fill-chars & Min-occ. & $\theta$ \\ 
\midrule
\multirow{5}{*}{True}
    & Constr.   &  0          & 7\enspace     & 0.5      &  & 1           & 11\enspace    & 0.35     \\
    & HurtLex-c &  0          & -\enspace     & -        &  & 0           & -\enspace     & -        \\
    & HurtLex-i &  0          & -\enspace     & -        &  & 0           & -\enspace     & -        \\
    & Wiegand-b &  0          & -\enspace     & -        &  & 0           & -\enspace     & -        \\
    & Wiegand-e &  0          & -\enspace     & -        &  & 0           & -\enspace     & -        \\ 
\midrule
\multirow{5}{*}{False}
    & Constr.   &  0          & 5\enspace     & 0.85     &  & 1           & 11\enspace    & 0.5      \\
    & HurtLex-c &  0          & -\enspace     & -        &  & 0           & -\enspace     & -        \\
    & HurtLex-i &  0          & -\enspace     & -        &  & 0           & -\enspace     & -        \\
    & Wiegand-b &  0          & -\enspace     & -        &  & 0           & -\enspace     & -        \\
    & Wiegand-e &  0          & -\enspace     & -        &  & 0           & -\enspace     & -        \\ 
\bottomrule
\end{tabular}
\end{subtable}

\begin{subtable}[htbp]{\textwidth}\centering

\caption{Rationale Extraction. Method-specific parameters are described in \autoref{sec:rationale-extraction-hyper-parameters}}

\setlength{\tabcolsep}{7.7pt}
\begin{tabular}{@{}llrrrrr@{}}
\toprule
\multirow{2}{*}{Prop-binary}
    &            & \multicolumn{2}{l}{HateXplain} &  & \multicolumn{2}{l}{SemEval}   \\ 
\cmidrule(r){3-4} \cmidrule(){5-7} 
    &            & Fill-chars & Threshold         &  & Fill-chars & Threshold        \\ 
\midrule
\multirow{4}{*}{True}
    & Saliency  \;& 0          & .055              &  & 0          & .081             \\
    & Int. Grad.\;& 0          & .081              &  & 1          & .133             \\
    & DeepLIFT  \;& 9999       & .133              &  & 0          & .133             \\
    & LIME      \;& 0          & .133              &  & 0          & .264             \\ 
\midrule
\multirow{4}{*}{False}
    & Saliency  \;& 0          & .107              &  & 1          & .133             \\
    & Int. Grad.\;& 0          & .238              &  & 0          & .474             \\
    & DeepLIFT  \;& 9999       & .290              &  & 0          & .317             \\
    & LIME      \;& 9999       & .500              &  & 0          & .500             \\ 
\bottomrule
\end{tabular}
\end{subtable}

\begin{subtable}[htbp]{\textwidth}\centering

\caption{Fine-tuned Language Models.}

\setlength{\tabcolsep}{7.7pt}
\begin{tabular}{@{}llrrr@{}}
\toprule
\multirow{2}{*}{Prop-binary}
    &          & \multicolumn{1}{l}{HateXplain} &  & \multicolumn{1}{l}{SemEval} \\ 
\cmidrule(r){3-3} \cmidrule(){5-5} 
    &          & Fill-chars    &                      & Fill-chars  \\ 
\midrule
\multirow{2}{*}{True}
    & BERT     &  0            &                      &  1           \\
    & BERT+CRF &  0            &                      &  1           \\ 
\midrule
\multirow{2}{*}{False}
    & BERT     &  0            &                      &  1           \\
    & BERT+CRF &  0            &                      &  1           \\ 
\bottomrule
\end{tabular}

\end{subtable}

\end{table} \fi

\end{document}